\documentclass{article}
\usepackage{ijcai21}

\usepackage{times}
\usepackage{soul}
\usepackage{url}
\usepackage[hidelinks]{hyperref}
\usepackage[utf8]{inputenc}
\usepackage[small]{caption}
\usepackage{graphicx}
\usepackage{amsmath}
\usepackage{amsthm}
\usepackage{booktabs}
\usepackage{algorithm}
\usepackage{algorithmic}
\usepackage{amsfonts} 
\DeclareMathOperator*{\argmax}{arg\,max}
\DeclareMathOperator*{\argmin}{arg\,min}
\usepackage{subfigure}
\usepackage{booktabs}
\usepackage{enumitem}
\usepackage{bbm}
\usepackage{lineno}

\title{Graph Universal Adversarial Attacks: \\
A Few Bad Actors Ruin Graph Learning Models}

\author{
Xiao Zang$^1$
\and
Yi Xie$^1$\and
Jie Chen$^2$\footnote{Contact Author}\And
Bo Yuan$^1$
\affiliations
$^1$Department of Electrical and Computer Engineering, Rutgers University\\
$^2$MIT-IBM Watson AI Lab, IBM Research
\emails
\{xz514, yx238\}@scarletmail.rutgers.edu,
chenjie@us.ibm.com,
 bo.yuan@soe.rutgers.edu
}

\begin{document}
\maketitle
\begin{abstract}
Deep neural networks, while generalize well, are known to be sensitive to small adversarial perturbations. This phenomenon poses severe security threat and calls for in-depth investigation of the robustness of deep learning models. With the emergence of neural networks for graph structured data, similar investigations are urged to understand their robustness. It has been found that adversarially perturbing the graph structure and/or node features may result in a significant degradation of the model performance. In this work, we show from a different angle that such fragility similarly occurs if the graph contains a few bad-actor nodes, which compromise a trained graph neural network through flipping the connections to any targeted victim. Worse, the bad actors found for one graph model severely compromise other models as well. We call the bad actors ``anchor nodes'' and propose an algorithm, named GUA, to identify them. Thorough empirical investigations suggest an interesting finding that the anchor nodes often belong to the same class; and they also corroborate the intuitive trade-off between the number of anchor nodes and the attack success rate. For the dataset Cora which contains 2708 nodes, as few as six anchor nodes will result in an attack success rate higher than 80\% for GCN and other three models.
\end{abstract}

\section{Introduction}\label{submission}
Graph structured data are ubiquitous with examples ranging from proteins, power grids, traffic networks, to social networks. Deep learning models for graphs, in particular, graph neural networks (GNN)~\cite{Kipf2017,Hamilton2017,velivckovic2017graph} attracted much attention recently and have achieved remarkable success in several tasks, including community detection, link prediction, and node classification. Their success is witnessed by many practical applications, such as content recommendation~\cite{wu2019session}, protein interaction~\cite{tsubaki2018compound}, and blog analysis~\cite{conover2011political}.

Deep learning models are known to be vulnerable and may suffer intentional attack with unnoticeable change of the data~\cite{zugner2018adversarial}. This observation originated from early findings by~\cite{Szegedy2014} and~\cite{goodfellow2014explaining}, who show that images perturbed with adversarially designed noise can be misclassified, while the perturbation is almost imperceptible.
This minor but intentional change would result in severe consequences socially and economically. For example, Wikipedia hoax articles can effectively disguise through modifying their links in a proper manner~\cite{kumar2016disinformation}. For another example, frauds may hide themselves through building plausible friendship in a social network, to confuse the prediction system.

In this work, we study the vulnerability of GNNs and show that it is possible to attack them if a few graph nodes serve as the bad actors: when their links to a certain node are flipped, the node will likely be misclassified. Such attacks are akin to universal attacks because the bad actors are universal to any target. We propose a graph universal adversarial attack method, GUA, to identify the bad actors.

Our work differs from recent studies on adversarial attack of GNNs~\cite{dai2018adversarial,zugner2018adversarial,jin2019power} in the attack setting. Prior work focuses on poisoning attacks (injecting or modifying training data as well as labels to foster a misbehaving model) and evasion attacks (modifying test data to encourage misclassification of a trained model). For graphs, these attacks could modify the graph structure and/or node features in a target-dependent scenario. The setting we consider, on the other hand, is a single and universal modification that applies to all targets. One clear advantage from the attack point of view is that computing the modification incurs a lower cost, as it is done once for all.
While universal attacks were studied earlier (see, e.g., \cite{moosavi2017universal} who compute a single perturbation applied to all images in ImageNet), graph universal attacks are rarely explored. This work contributes to the literature a setting and a method that may inspire further study on defense mechanisms of deep graph models.

\begin{figure}[t]
  \centering
  \includegraphics[width=\linewidth]{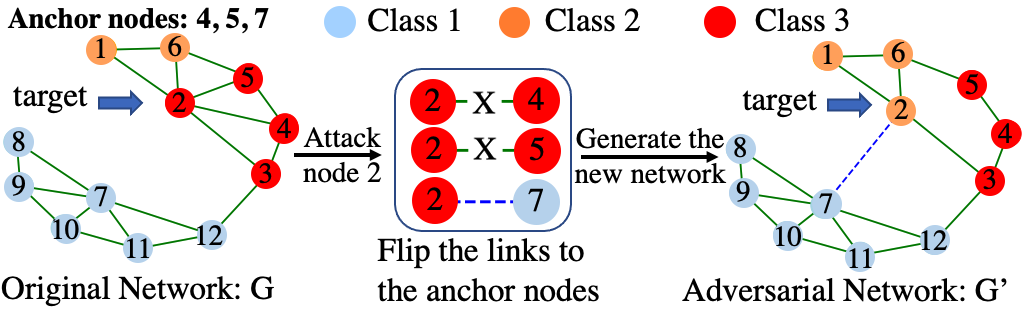}
  \vskip-5pt
  \caption{Illustration of GUA. A small number of anchor nodes (4, 5, and 7) is identified. To confuse the classification of a target node (e.g., 2), their connections to this node are flipped.}
  \label{framework_fg}
\end{figure}

Figure~\ref{framework_fg} illustrates the universal attack setting we consider. A few bad-actor nodes (4, 5, and 7) are identified; we call them \emph{anchor nodes}. When an adversary attempts to attack the classification of a target node (say, 2), the existing links from the anchor nodes to the target node are removed while non-existing links are created. The identification method we propose, GUA, is conducted on a particular classification model (here, GCN), but the found anchors apply to other models as well (e.g., DeepWalk, node2vec, and GAT).

As a type of attacks, universal attacks may be preferred by the adversary for several reasons. First, the anchor nodes are computed only once and there incurs no extra cost when attacking individual targets. Second, the number of anchors can be very small (it is easier to compromise fewer nodes). Third, attacks are less noticeable when only a limited number of links are flipped.


The contribution of this work is threefold:
\begin{itemize}[leftmargin=*]
  
\item We propose a novel algorithm for graph universal attack that achieves high success rate and demonstrates vulnerability of graph deep learning models.
  
\item We demonstrate appealing generalization of the attack algorithm, which finds anchor nodes based on a small training set but successfully attacks a majority of target nodes.
  
\item We show attractive transferability of the found anchors (based on GCN) through demonstrating similar attack success rates on other graph learning models.
\end{itemize}

\section{Notation and Background}
A graph is denoted as $G=(V,E)$, where $V$ is the node set and $E$ is the edge set. An unweighted graph is represented by the adjacency matrix $A=\{0,1\}^{|V|\times|V|}$.
The graph nodes have $d$-dimensional features, which collectively form the feature matrix $X$, whose dimension is $|V|\times d$.

In GCN~\cite{Kipf2017}, one normalizes the adjacency matrix into $\widehat{A}=\widetilde{D}^{-\frac{1}{2}}\widetilde{A}\widetilde{D}^{-\frac{1}{2}}$, where $\widetilde{A} = A + I$ and $\widetilde{D}$ is the diagonal adjusted degree matrix with diagonal entries $\widetilde{D}_{ii}=\sum_{j=1}^{|V|}\widetilde{A}_{ij}$. Then, the neural network is
\begin{equation}
Z = f(A,X) = \text{softmax}(\hat{A}\text{ReLU}(\hat{A}XW^{(0)})W^{(1)}),
\end{equation}
where $W^{(0)}$ and $W^{(1)}$ are model parameters. The training of the parameters uses the cross-entropy loss.

\section{Graph Universal Adversarial Attack}
Following the notation introduced in the preceding section, given the graph adjacency matrix $A$ and node feature matrix $X$, we let $f(A,X)$ be the classification model and let $\hat{l}(A,X,i)$ be the predicted label of node $i$; that is, $\hat{l}(A, X, i) = \argmax f(A, X)_i$.

Given a trained model $f$, the goal is for each node $i$ to modify the adjacency matrix $A$ into $A'$ such that $\hat{l}(A', X, i) \neq \hat{l}(A, X, i)$.
Note that the modified $A'$ is $i$-dependent in our attack setting.

\subsection{Attack Vector and Matrix}
Let the graph have $n$ nodes. We use a length-$n$ binary vector $p$ to denote the attack vector to be determined, where 1 means an anchor node and 0 otherwise. Hence, $A'$ is a function of three quantities: the original adjacency matrix $A$, the target node $i$, and the attack vector $p$.

To derive an explicit form of the function, we extend the vector $p$ to an $n\times n$ matrix $P$, which starts as a zero matrix, with the $i$th row and $i$th column replaced by the attack vector $p$. Thus, the $(i,j)$ element of the attack matrix $P$ indicates whether the connection of the node pair $(i,j)$ is flipped: 1 means yes and 0 means no.

It is then not hard to see that one may write the function
\begin{equation}\label{perturb_eq}
  A' := g(A,i,p) = (\mathbf{1} - P) \circ A + P \circ (\mathbf{1}_0 - A),
\end{equation}
where $\mathbf{1}$ denotes the matrix of all ones and $\mathbf{1}_0$ is similar except that the diagonal is replaced by zero; $\circ$ means element-wise multiplication. The term $(\mathbf{1} - P)$ serves as the mask that preserves the connections of all node pairs other than those between the anchors $j$ and the target node $i$. The term $(\mathbf{1}_0 - A)$ intends to flip the whole $A$ (except diagonal) but the $P$ in the front ensures that only the involved $(i,j)$ pairs are actually flipped. Moreover, one can verify that the diagonal of the new adjacency matrix remains zero.

For gradient based optimization, the binary elements of the attack vector $p$ may be relaxed into real values between 0 and 1. In this case, the connections of all node pairs other than those between the anchors $j$ and the target node $i$ remain the same. On the other hand, the connections between the involved $(i,j)$ pairs are fractionally changed. The $j$th element of $p$ indicates the strength of change.

\subsection{Outer Procedure: GUA}
Let $V_L$ be the training set with known node labels. Given an attack success rate threshold $\delta$, we formulate the problem as finding a binary vector $p$ such that
\begin{equation}\label{eqn:Err}
  Err(V_L) := \frac{1}{|V_L|} \sum\limits_{i=1}^{|V_L|}\mathbf{1}\{\hat{l}(A', X, i) \neq \hat{l}(A, X, i)\} \geq \delta.
\end{equation}

To effectively leverage gradient-based tools for adversarial attacks, we perform a continuous relaxation on $p$ so that it can be iteratively updated. Now elements of $p$ stay in the interval $[0,1]$. The algorithm proceeds as follows. We initialize $p$ with zero. In each epoch, we begin with a binary $p$ and iteratively visit each training node $i$. If $i$ is not misclassified by the current $p$, we seek a minimum continuous perturbation $\Delta p$ to misclassify it. In other words,
\begin{equation}\label{eqn:Delta.p}
\Delta p = \argmin_r ||r||_2 ,\,\,\,
\text{s.t.} \,\,\, \hat{l}(g(A, i, p+r), X, i) \neq \hat{l}(A, X, i).
\end{equation}
We will elaborate in the next subsection an algorithm to find such $\Delta p$. After all training nodes are visited, we perform a hard threasholding at 0.5 and force back $p$ to be a binary vector. Then, the next epoch begins. We run a maximum number of epochs and terminate when~\eqref{eqn:Err} is satisfied.

The updated $p \gets p+\Delta p$ found through solving~\eqref{eqn:Delta.p}, if unbounded, may be problematic because (i) it may incur too many anchor nodes and (ii) its elements may be outside $[0,1]$. We perform an $L_2$-norm projection to circumvent the first problem and a clipping between interval [0, 1] to circumvent the second. The rationale of $L_2$-norm projection is to suppress the magnitude of $p$ and encourage that eventually few entries are greater than 0.5. The maximum number of anchor nodes grows quadratically with the projection radius $\xi$. A small $\xi$ clearly encourages fewer anchors.

Through experimentation, we find that clipping is crucial to obtaining a stable result. In a later section, we illustrate an experiment to show that the attack success rate may drop to zero in several random trials, if clipping is not performed. See Figure~\ref{comp_fg}.

The procedure presented so far is summarized in Algorithm~\ref{graphuniversalattack_alg}. The algorithm for obtaining $\Delta p$ is called IMP (iterative minimum perturbation) and will be discussed next.

\begin{algorithm}[t]
\small
  \caption{Graph Universal Attack (GUA)}
  \label{graphuniversalattack_alg}
  \begin{algorithmic}
    \STATE \textbf{Input:} adjacency matrix $A$, node features $X$, $max\_epoch$, max iteration $max\_iter$, $\delta$, data $overshoot$
    \STATE $p \leftarrow 0$
    \WHILE{$epoch < max\_epoch$}
    \FOR{$i \in V_L$}
    \STATE $A' \leftarrow g(A,i,p)$
    \IF{$\hat{l}(A', X, i) = \hat{l}(A, X, i)$}
    \STATE $\Delta p, iter \leftarrow \text{IMP}(A', i, overshoot, max\_iter)$
    \STATE $p \leftarrow p+\Delta p$
    \STATE $p \leftarrow \text{$L_2$-norm projection}(p)$
    \STATE $p \leftarrow p.clip(0,1)$
    \ENDIF
    \ENDFOR
    \STATE $p \leftarrow (p>0.5)\,?\,1:0$
    \STATE compute $Err(V_L)$ 
    \STATE \textbf{if} $Err(V_L) \ge \delta$, \textbf{break}
    \ENDWHILE
    \STATE \textbf{return} $p$
  \end{algorithmic}
\end{algorithm}


\subsection{Inner Procedure: IMP}
To solve~\eqref{eqn:Delta.p}, we adapt DeepFool~\cite{moosavi2016deepfool} to find a minimum perturbation that sends the target node $i$ to the decision boundary of another class.

Denote by $v$ the minimum perturbation. To find the closest decision boundary other than that of the original class $pred = \hat{l}(A, X, i)$, we first select the closest class $k = \argmin_{c \neq pred} \frac{|\Delta f_c|}{\|\Delta w_c\|_2}$,
where $\Delta f_c = f(A, X)_{i,c} - f(A, X)_{i,pred}$ and $\Delta w_c = \nabla f(A, X)_{i,c} - \nabla f(A, X)_{i, pred}$. Then, we update $v$ by adding to it $\Delta v$:
\begin{equation}
  \Delta v = \frac{|\Delta f_{k}|}{||\Delta w_{k}||_2^2} \Delta w_{k}.
\label{deepfool_eq}
\end{equation}
We iteratively update $v$ until $(1+overshoot) \times v$ successfully attacks node $i$, where $overshoot$ is a small factor that ensures the node passes the decision boundary. We also clip the new $A'$ to ensure stability, in a manner similar to the handling of $p$ in the preceding subsection.
The procedure for computing the minimum perturbation $v$ is summarized in Algorithm~\ref{deepfool_alg}.

\begin{algorithm}[h]
\small
  \caption{Iterative Minimum Perturbation (IMP)}
  \label{deepfool_alg}
  \begin{algorithmic}
    \STATE \textbf{Input:} $A$, node index $i$, $overshoot$, $max\_iter$
    \STATE $v \leftarrow 0$
    \STATE $iter \leftarrow 0$
    \STATE $pred \leftarrow \hat{l}(A, X, i)$
    \STATE $A' \leftarrow A$
    \WHILE{$\hat{l}(A', X, i) = pred$ \text{and} $iter < max\_iter$}
    \STATE $\Delta v \leftarrow \frac{|\Delta f_k|}{||\Delta w_k||_2^2} \Delta w_k \,\,\,\text{according to Equation ~\eqref{deepfool_eq}}$
    \STATE $v \leftarrow v + \Delta v$
    \STATE $A' \leftarrow g(A, i, (1+overshoot) \times v).clip(0, 1)$
    \ENDWHILE
    \STATE $v \leftarrow (1+overshoot) \times v$
    \STATE \textbf{return} $v, iter$
  \end{algorithmic}
\end{algorithm}
\section{Experiments}\label{exp_sec}
In this section, we evaluate thoroughly the proposed attack GUA through investigating its design details, comparing with baselines, performing scalability tests, and validating transferability from model to model. Code is available at \url{https://github.com/chisam0217/Graph-Universal-Attack}.

\subsection{Datasets and Settings}
We compute the anchor set through attacking the standard GCN model. The parameters of Algorithms~\ref{graphuniversalattack_alg} and~\ref{deepfool_alg} are: $max\_epoch=100$, $max\_iter=20$, $\delta=0.9$, and $overshoot=0.02$. To cope with randomness, experiments are repeated ten times for each setting. We work with three commonly used node classification benchmark datasets~\cite{sen2008collective}. Their information is summarized in Table~\ref{dataset_tb}.

\begin{table}[b]
\small
  \centering
  \begin{tabular}{lccc}
    \toprule
    Statistics & Cora & Citeseer & Pol.Blogs \\ 
    \midrule
    Nodes(LCC) & $2708$ & $3327$ & $1222$ \\
    Edges(LCC) & $5278$ & $4676$ & $16714$ \\
    Classes & $7$ & $6$ & $2$ \\
    Train/test split & $140/1000$ & $120/1000$ & $121/1101$ \\
    Accuracy(GCN) & $81.4\%$ & $70.4\%$ & $94.3\%$ \\
    \bottomrule
    \vspace{-5mm}
  \end{tabular}
  \caption {Dataset Statistics. Only the largest connected component (LCC) is considered.}
  \label{dataset_tb}
  
\end{table}

\subsection{Baseline Methods}\label{sec:baseline}
Because graph universal attacks were barely studied, we design four relavant baselines for GUA. The first two are basic while the last two are more sophisticated.
\begin{itemize}[leftmargin=*]
\item Global Random: Each node has a probability $Prob$ to become an anchor node. In other words, each element of the attack vector $p$ is an independent sample of Bernoulli($Prob$). 
\item Victim-Class Attack (Victim Attack): We sample a prescribed number of anchor nodes without replacement from nodes of a particular class. This baseline originates from a finding that the anchor nodes computed by GUA often belong to the same class (see more details later).
\item High-Degree (HD) Global Random: We strengthen the Global Random baseline by picking random anchors uniformly from top 10\% nodes with highest degrees.
\item Top-Confidence (TC) Victim Attack: The anchor set is composed of nodes with the highest prediction probability from the victim class.

\end{itemize}

Additionally, we compare with Fast Gradient Attack (FGA)~\cite{chen2018fast} and Nettack~\cite{zugner2018adversarial}, both of which are per-node attack methods. They are not universal attacks. FGA flips edges/non-edges connecting to different nodes depending on the target, while Nettack modifies not only the edges, but also the node features. We also compare with Meta-Self Attack~\cite{zugner2019adversarial}, which performs the global attack by perturbing the graph structure through meta learning.

\subsection{Results}\label{results}
The evaluation metric is attack success rate (ASR). Another quantity of interest is the number of modified links (ML). For universal attacks, it is equivalent to the anchor set size.

\begin{figure}[b]
    \centering
    \subfigure[Clipping.]{
    \includegraphics[height=1.55in]{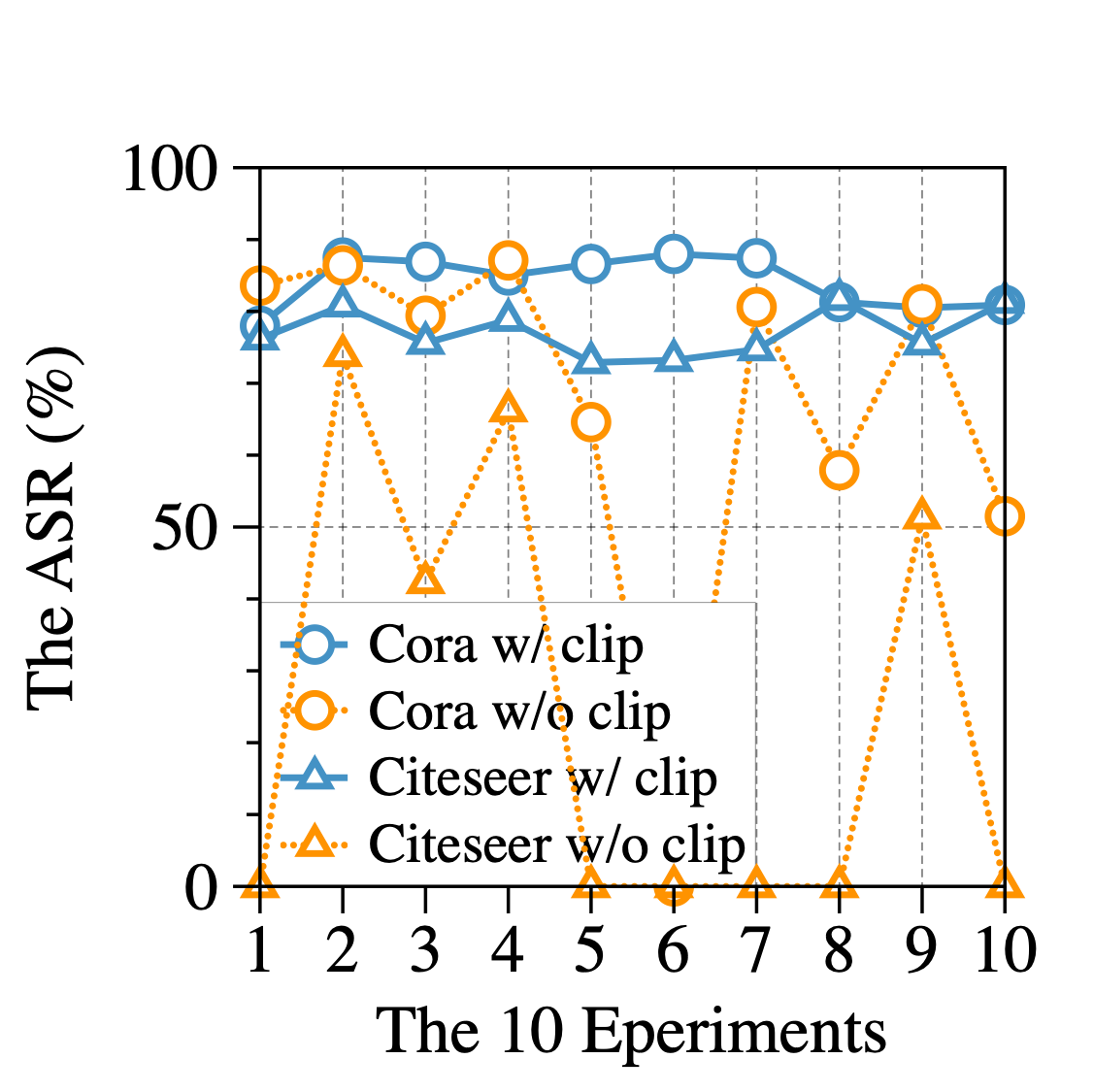} 
    \label{comp_fg}}
    \subfigure[Deletion of anchor nodes.]{
    \includegraphics[height=1.55in]{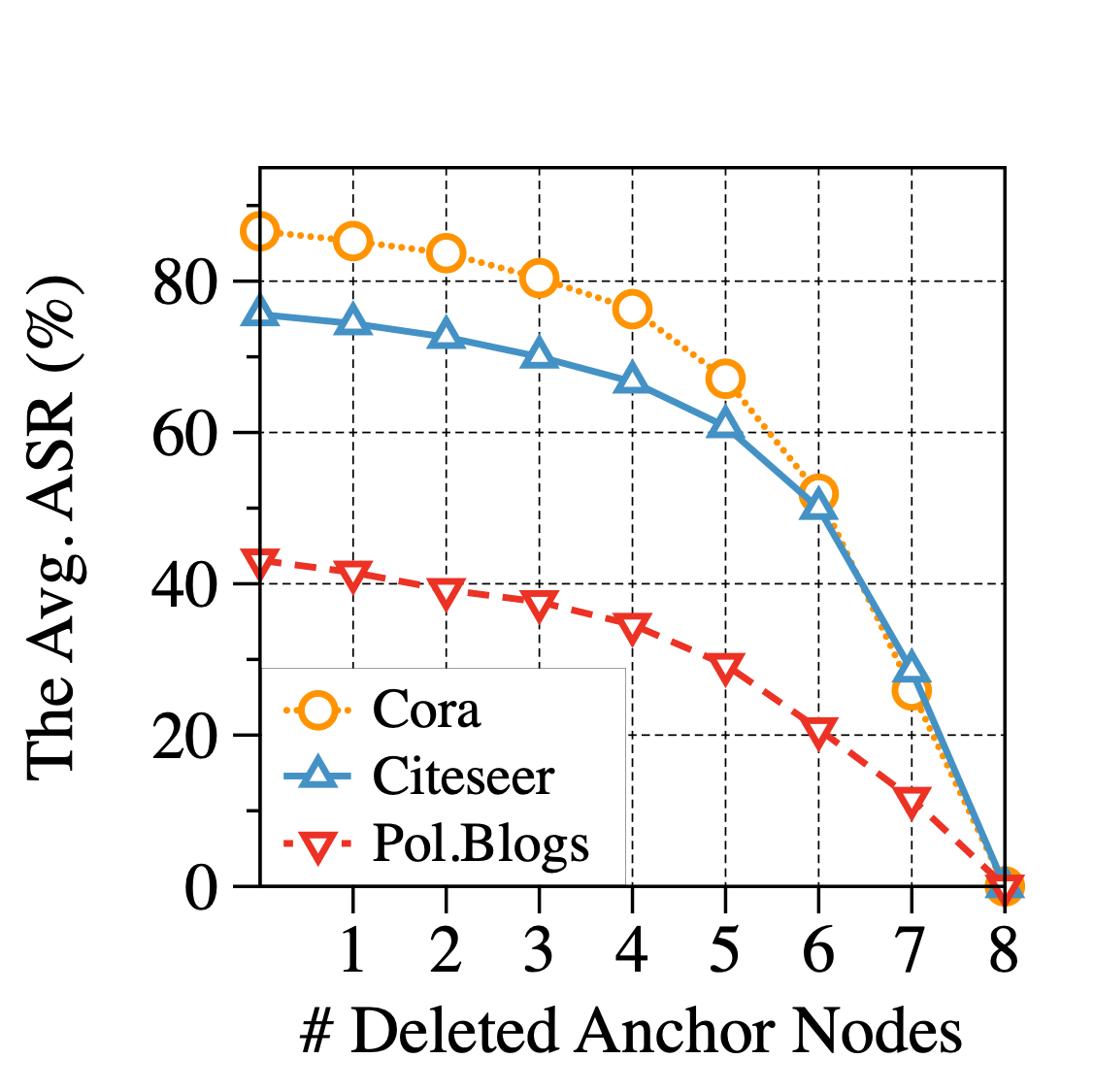}
    \label{universaldelete_fg}}
    \caption{Left: ASRs of using clipping versus not. Right: average ASR after deleting nodes from the anchor set.}
    \label{fig:my_label}
\end{figure}

\paragraph{Importance of clipping.}
As discussed in the design of GUA, the continuous relaxation of the attack vector $p$ requires clipping throughout optimization. For an empirical supporting evidence, we show in Figure~\ref{comp_fg} the ASR obtained through executing Algorithm~\ref{graphuniversalattack_alg} with and without clipping, respectively. Clearly, clipping leads to stabler and superior results. Without clipping, the ASR may drop to zero in some random trials. The reason is that several entries of $p$ become strongly negative, such that projections result in small values for all positive entries and subsequent hard thresholding zeros out the whole vector $p$. 

\paragraph{Effect of projection radius.}
The $L_2$-norm projection radius is a positive quantity $\xi$ so that the projection of $p$ is $\xi\cdot p/\|p\|_2$. The number of anchors is implicitly controlled by the projection radius $\xi$. Increasing $\xi$ will enlarge the anchor set. We treat $\xi$ as a parameter and study its relationship with the number of anchor nodes and corresponding ASR.  See Table~\ref{data_tb}. As expected, the average ML (that is equal to the number of anchor nodes) increases with $\xi$ non-linearly. A larger anchor set also frequently results in higher ASR, because of more changes to the graph.
\begin{table*}[ht]
    \centering
    \small
    \begin{tabular}{lcccccccccc}
    \toprule
    \multicolumn{1}{l}{} & \multicolumn{3}{c}{Cora} & \multicolumn{3}{c}{Citeseer} & \multicolumn{4}{c}{Pol.Blogs} \\
    \cmidrule(r){2-4}
    \cmidrule(r){5-7}
    \cmidrule(r){8-11}
    & $\xi =3$ & $\xi=4$ & $\xi=5$ &$\xi =3$ & $\xi=4$ & $\xi=5$ &$\xi =3$ & $\xi=4$ & $\xi=5$ & $\xi=8$\\
    \midrule
    Avg. ML & $4.4$ & $7.9$ & $10.8$ & $3.3$ & $7.7$ & $13.9$ & $2.6$ & $5.3$ & $10$ & $26.6$\\
    Avg. ASR(\%) & $75.60$ & $84.21$ & $83.56$ & $63.67$ & $77.07$ & $80.01$ & $17.68$ & $22.65$ & $30.84$ & $43.06$ \\
    \bottomrule
    \vspace{-5mm}
    \end{tabular}
    \caption{Average ML and average ASR under different projection radius $\xi$.}
    \label{data_tb}
    \vspace{-5mm}
\end{table*}
\begin{figure}
    \centering
    \includegraphics[width=3.2in]{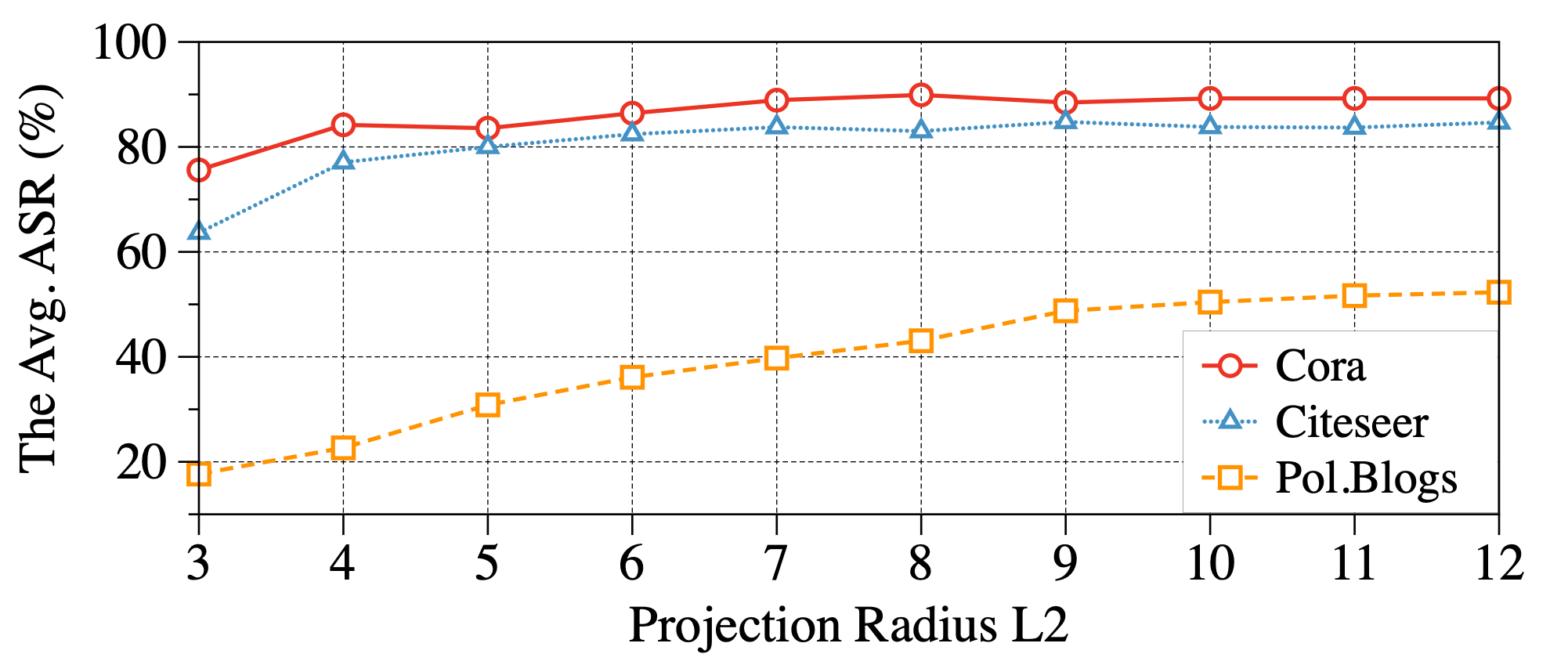}
    \caption{The average ASRs on all datasets with respect to projection radius $L_2$.}
    \label{fig:radius_all}
\end{figure}
Additionally, the individual result for each trial may suggest even more attractive findings. For example, for the case of Cora and $\xi=4$, the MLs for the ten trials are \{5, 9, 10, 7, 8, 9, 9, 9, 6, 7\} and the corresponding ASRs are \{0.780, 0.875, 0.869, 0.850, 0.866, 0.880, 0.874, 0.813, 0.805, 0.809\}. This result means that as few as six anchor nodes are sufficient to achieve 80\% ASR. The different number of anchors under the same $\xi$ also corroborates the above statement that the number of anchors is implicitly controlled by the projection radius $\xi$. 
In Figure~\ref{fig:radius_all}, we plot the average ASRs with respect to different $L_2$-norm projection radius. For Cora, Citeseer and Pol.Blogs, the plateaued ASR is $89.91\%$, $84.77\%$ and $52.31\%$, at $\xi = 8$, $\xi = 9$ and $\xi = 12$, respectively.
On the other hand, one also sees that the attacks on Pol.Blogs are less effective. The reason is that the graph has a large average degree, which makes it relatively robust to universal attacks. As observed by~\cite{zugner2018adversarial} and~\cite{wu2019adversarial}, nodes with more neighbors are harder to attack than those with fewer neighbors. The higher density of the graph requires a larger anchor set to achieve high ASR. 
We intend to maintain a small anchor set for secretive attacks.
Thus, keeping the other hyperparameters the same, we report future results by setting $\xi = 4, 4, 8$ on Cora, Citeseer and Pol.Blogs, respectively.

\paragraph{Nodes with low connectivity are easier to attack.} On Cora, when $\xi = 4$, the average degree of successfully attacked nodes is $4.29$, while that of failed ones is $7.08$. This result also corroborates the conclusions from ~\cite{zugner2018adversarial}, that low-degree nodes are easier to attack. The phenomenon is not surprising since the neighborhoods of low-degree nodes tend to be dominated by anchors.  

\begin{figure}[t]
  \centering
  \subfigure[Cora.]{
    \includegraphics[height=1.6in]{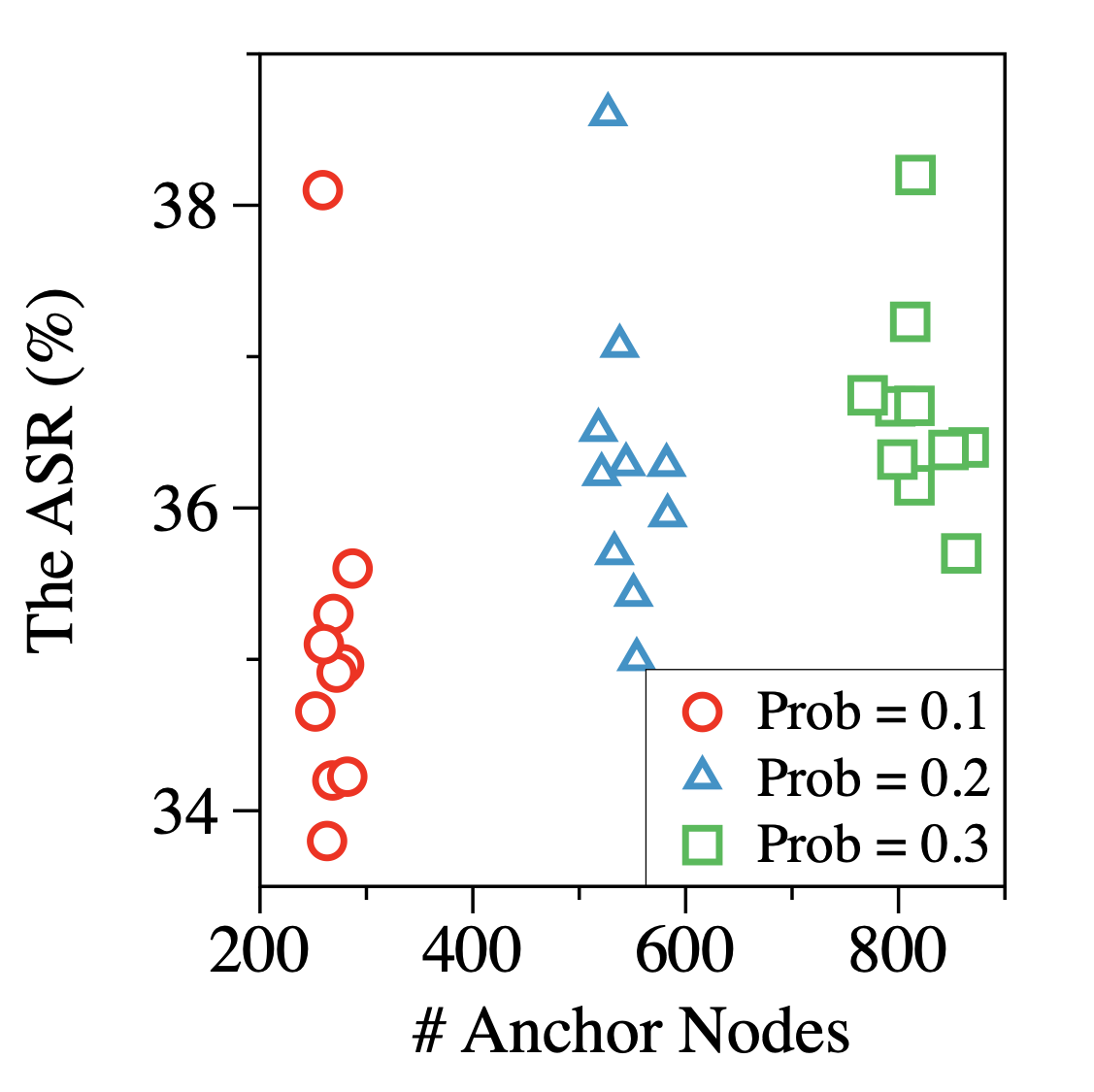}}
  \subfigure[Citeseer.]{
    \includegraphics[height=1.6in]{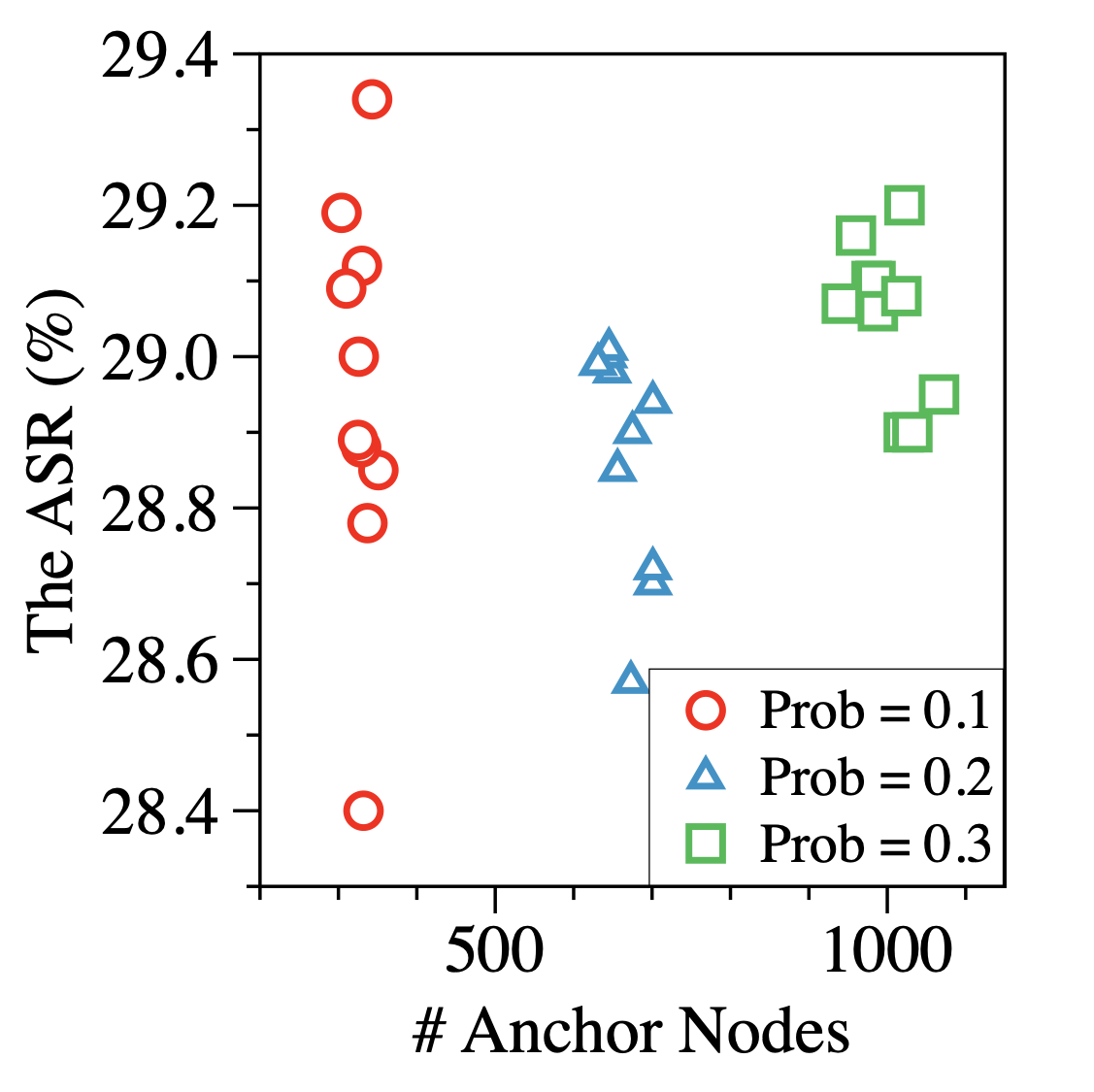}}
  \caption{Performance of global random attack, repeated ten times.}
  \label{fig:globalrandom}
\end{figure}

\paragraph{Blindly using more anchors does not work.}
Now that we have an effective method to compute the anchor nodes, we investigate whether randomly chosen anchor nodes are similarly effective. In Figure~\ref{fig:globalrandom}, we plot the ASR results of Global Random. One sees that all ASRs for Cora are below 40\% and for Citeseer below 30\%. Such results are way inferior to those of GUA. Moreover, using hundreds and even a thousand anchors does not improve the ASR. 

\paragraph{Anchors often belong to the same class.}
With analysis of the anchors, an interesting finding is that one class dominates. The average entropies of the anchors' class distribution on Cora, Citeseer and Pol.blogs are $0.1$, $0.1$, and $0$, respectively. This indicates that in most of the cases only one class appears. One possible reason is that anchors united by the same class form an overwhelming influence: either to allure a node into this class through establishing links or kick it out of the class through eliminating links. Actually, one of our baselines, Victim-Class Attack, is designed for such a phenomenon.

\paragraph{Wrong classifications often coincide with the anchor class.}
A natural conjecture following the above finding is that a target node will be misclassified to the (majority) class of the anchors. Our experiment on Cora corroborates this conjecture. The results indicate that 96\% of the test nodes are misclassified to class 6 when all the anchor nodes belong to this class. An analysis of the dataset shows that each node has two neighbors on average. Hence, flipping the connections to the anchor nodes possibly makes the anchor class dominate among the new set of neighbors. Then, classifying into the anchor class becomes more likely. This result echoes one mentioned by~\cite{nandanwar2016structural}, who conclude that classification of a node is strongly influenced by the classes of its neighbors; it tends to coincide with the majority class of the neighbors.
To generalize the above result, we collect the entropy change of the class distribution before and after attack. On Cora, Citeseer and Pol.Blogs, the changes are respectively $-1.66$, $-1.49$ and $-0.07$.
For all datasets, the entropy decreases, indicating stronger dominance of one class after attack. The decrease is more substantial for Cora and Citeseer than for Pol.Blogs, which is expected, because the latter has denser and more varied connections, which eclipse the dominance of the anchor class.
To further illustrate the dominance of the anchor class, we investigate the ASRs of the nodes from anchor class versus non-anchor class. The ASRs are strikingly different. For example, on Cora, when $\xi = 4$, the ASR of nodes from the “anchor class” is $0\%$, while that of nodes from “non-anchor class” is $96.5\%$. Such a phenomenon is expected. In most cases, flipping the edges with the anchor nodes results in dominance of the anchor class. Furthermore, we find that $99.9\%$ of the successfully attacked targets are misclassified into the anchor class, corroborating the theory.

\paragraph{Comparison with baselines.}
We compare GUA with four baseline methods explained in Section~\ref{sec:baseline}, together with two non-universal attacks and one global attack. Since we cannot choose the number of anchor nodes for GUA, we obtain this value based on the results in Table~\ref{dataset_tb} when $\xi=4, 4, 8$ on Cora, Citeseer and Pol.Blogs, respectively. In this case, the average ML for these datasets is respectively 7.9, 7.7, and 26.6. Therefore, we set the number of anchor nodes for all baselines, but for Global Random and Meta-Self, to be the ceiling of these values. For Global Random, $Prob$ is set such that the expected number of anchor nodes is these values. For Meta-Self, the MLs are 53, 47 and 167 for Cora, Citeseer and Pol.Blogs, respectively, which are $1\%$ of the number of original edges in corresponding datasets.
\begin{table}[b]
\small
  \centering
  \begin{tabular}{lcccr}
    \toprule
    Attack Method & Cora & Citeseer & Pol.Blogs \\ 
    \midrule
    GUA & $84.21\%$ & $77.07\%$ & $43.06\%$ \\
    Global Random & $22.00\%$ & $26.58\%$ & $17.58\%$ \\
    Victim Attack & $62.64\%$ & $54.47\%$ & $32.45\%$ \\
    HD Global Random & $26.68\%$ & $51.74\%$ & $17.28\%$ \\
    TC Victim Attack & $79.64\%$ & $73.45\%$ & $36.09\%$ \\
    FGA (not univ.) & $89.70\%$ & $84.82\%$ & $57.67\%$ \\
    Nettack (not univ.) & $86.09\%$ & $77.06\%$ & $76.91\%$ \\
    Meta-Self (global) & $16.21\%$ & $30.37\%$ & $14.25\%$ \\
    \bottomrule
    \vspace{-5mm}
  \end{tabular}
  \caption {Average ASR. For a fair comparison, all universal attack methods except Global Random use the same number of anchor nodes. FGA and Nettack are not universal attacks and we set their ML per node to be the same as the number of anchor nodes.}
  \label{baseline_tb}
\end{table}
From Table~\ref{baseline_tb}, one sees that GUA significantly outperforms other universal attack methods. Among them, Top-Confidence Victim-Class Attack is the most effective, but it is still inferior to GUA. This result suggests that GUA leverages more information (in this case, node features) than the class labels, although we have seen strong evidences that anchor nodes computed by GUA mostly belong to the same class.
One also sees that GUA is inferior to FGA and Nettack if only ASR is concerned. FGA and Nettack are not universal attack methods; they find different anchors for each target node. Thus, it is possible to optimize the number of anchors (possibly different for each target) to aim at a certain ASR, or equivalently, to achieve a better ASR given a certain number of anchors. However, it is also because they are non-universal attacks, that the total number of anchors for all targets soars. For example, FGA modifies links with 1406 anchors on Cora and 1359 anchors on Citeseer in total. GUA also significantly outperforms Meta-Self, since it only attacks the graph once, instead of attacking each node individually.

\paragraph{Effect of removing anchor nodes.}
Once a set of anchor nodes is identified, a natural question asks if the set contains redundancy. We perform the following experiment: we randomly remove a number of anchor nodes and recompute the ASR. Because on Cora and Citeseer, the average number of anchor nodes for $\xi=4$ is 7.9 and 7.7 respectively, we use an anchor set of size eight to conduct the experiment. For each case, we randomly remove 1--7 nodes from the anchor set and report the corresponding average ASR. The results are shown in Figure~\ref{universaldelete_fg}. From the figure, one sees that the average ASR gradually decreases to zero as more and more anchor nodes are removed. This result indicates that there exists no redundancy in the anchor set. The decease is faster when more nodes are removed, but the average ASR is still quite high even when removing half of the nodes. This finding is another evidence that supports the trade-off between anchor set size and ASR, in addition to the prior Table~\ref{data_tb}.

\paragraph{Speeding up training through sampling.}
The cost of finding the anchors is $O(n \cdot |V_L| \cdot max\_epoch)$, because in each epoch the attack vector $p$ is kept being updated through iterating $|V_L|$ training nodes. For a given graph with fixed $n$, we are interested in seeing whether reducing the training set size affects the attack performance. We randomly sample a portion of the training set in each epoch and report in Table~\ref{sample_tb} the resulting ASR. One sees that the ASR barely changes by using 40\% of the data to train each epoch. Further reducing the size starts to hurt, but even using 5\% of the training data, the ASR drops by only 5\% to 13\%. 
Moreover, GUA is efficient compared to the state-of-art attack method Nettack~\cite{zugner2018adversarial}, whose complexity is $O(n^2 \cdot (E \cdot T+F))$ to attack all nodes, where $E$ and $F$ represent the number of edge and feature perturbations, respectively, and $T$ is the average size of a 2-hop neighborhood. In practice, Nettack is slower to run, due to the $n^2$ factor.

\begin{table}[t]
\vspace{-2mm}
\centering
\small
\begin{tabular}{lccccc}
\toprule
Dataset & $100\%$ & $40\%$ & $20\%$ & $10\%$ & $5\%$ \\
\midrule
Cora & $84.2\%$ & $83.7\%$ & $83.2\%$ & $82.6\%$ & $79.9\%$ \\
Citeseer & $77.1\%$ & $77.4\%$ & $69.7\%$ & $69.7$ & $65.2\%$ \\
Pol.Blogs & $43.1\%$ & $40.1\%$ & $35.7\%$ & $37.7\%$ & $30.8\%$ \\
\bottomrule
\vspace{-5mm}
\end{tabular}
\caption{Average ASR through sampling the training set per epoch.}
\label{sample_tb}
\end{table}

\paragraph{Transferability.}
We have already seen that GUA is quite effective in attacking GCN.
Such an attack belongs to the white-box family, because knowledge of the model to be attacked is assumed.
In reality, however, the model parameters may not be known at all, not even the model form. Attacks under this scenario is called black-box. One approach to conducting black-box attack is to use a surrogate model. In our case, if one is interested in attacking graph deep learning models other than GCN, GCN may serve as the surrogate. The important question is whether anchors found by attacking the surrogate can effectively attack other models as well.
We perform an experiment with three such models: DeepWalk~\cite{perozzi2014deepwalk}, node2vec~\cite{grover2016node2vec}, and GAT~\cite{velivckovic2017graph}. The first two compute, in an unsupervised manner, node embeddings that are used for downstream classification, whereas the last one is a graph neural network that directly performs classification. In Table~\ref{transer_tb}, we list the ASR for these models. One sees that the ASRs are similarly high as that for GCN; sometimes even surpassing. Specifically, GAT is developed based on GCN through incorporating the attention mechanism, while Node2vec and DeepWalk update node embeddings by exploring the local structure via random walk. Since GUA modifies the neighborhood of the target, it is reasonable that all the other methods can also be misled efficiently. This finding concludes that the results of GUA are well transferable.

\begin{table}[t]
\vspace{-2mm}
\centering
\small
\begin{tabular}{lcccc}
\toprule
Dataset & GCN & DeepWalk & node2vec & GAT \\ 
\midrule
Cora &$84.21\%$ &$85.80\%$ &$80.84\%$ &$85.15\%$ \\
Citeseer &$77.07\%$ &$81.71\%$ &$74.07\%$ &$77.02\%$\\
Pol.Blogs & $43.06\%$ & $33.21\%$ & $41.62\%$ & $38.85\%$ \\ 
\bottomrule
\vspace{-5mm}
\end{tabular}
\caption {Average ASR after applying the anchor nodes found by GUA when $\xi = 4$ on Cora and Citeseer, and $\xi = 8$ on Pol.Blogs.}
\label{transer_tb}
\end{table}

\section{Related Work}
Since the seminal work by~\cite{Szegedy2014}, various types of methods have been proposed to generate adversarial examples. For instance, \cite{goodfellow2014explaining} introduce the fast gradient sign method and \cite{carlini2017towards} develop a powerful attack through iterative optimization. This work is related to recent advances in adversarial attacks on GNNs and general universal attacks.

\paragraph{Adversarial attack on GNN.} \cite{zugner2018adversarial} propose Nettack that uses a greedy mechanism to attack the graph embedding model by changing the entry that maximizes the loss change. \cite{dai2018adversarial} introduce a reinforcement learning based method that modifies the graph structure and significantly lowers the test accuracy. \cite{wang2018attack} propose Greedy-GAN that poisons the training nodes through adding fake nodes indistinguishable by a discriminator. \cite{chen2018fast} recursively compute connection-based gradients and flip the connection value based on the maximum gradient. \cite{zugner2019adversarial} attack the whole graph by tackling an optimization problem using meta learning.

\paragraph{Universal attacks.} \cite{moosavi2017universal} train a quasi-imperceptible universal perturbation that successfully attacks most of the images in the same dataset. \cite{46561} generate a small universal patch to misclassify any image. \cite{wu2019universal} search for a universal perturbation that can be well transferred to several models.

\section{Conclusion}
In this work, we consider universal adversarial attacks on graphs and propose the first algorithm, named GUA, to effectively conduct these attacks. GUA finds a set of anchor nodes to mislead the classification of all nodes in the graph through flipping the connections between the anchors and the target node. GUA achieves the highest ASR compared to several universal attack baselines. The training process can be accelerated through sampling the training set in each epoch. There exists a trade-off between ASR and the anchor set size and we find that a very small size is sufficient to achieve remarkable attack success. Additionally, we find that the computed anchor nodes often belong to the same class. We also find that the anchor nodes used to attack one model equally well attack other models. In the future, we plan to develop defense mechanisms to effectively counter these attacks.

\section*{Acknowledgements}
J. Chen is supported in part by DOE Award DE-OE0000910.

\bibliographystyle{named}
\bibliography{reference}

\end{document}